\documentclass[letterpaper, 10 pt, conference]{ieeeconf}  

\IEEEoverridecommandlockouts                              

\usepackage{graphics} 
\usepackage{epsfig} 
\usepackage{times} 
\usepackage{amsmath} 
\usepackage{mathrsfs}
\usepackage{bm}
\usepackage{stfloats}
\usepackage{amssymb}  

\usepackage{subfigure}
\usepackage[T1]{fontenc}
\usepackage[vlined,ruled]{algorithm2e}
\usepackage{booktabs}
\usepackage{multirow}

\usepackage{graphicx}
\usepackage{float}

\usepackage{cite}
\usepackage{xcolor}
\usepackage{url}
\makeatletter
\let\NAT@parse\undefined
\makeatother
\usepackage{hyperref}
\usepackage{algpseudocode}
\usepackage[export]{adjustbox}
\usepackage{balance}
\usepackage[utf8]{inputenc}
\usepackage{verbatim}
\usepackage{makecell}
\usepackage{multicol}

\renewcommand{\theenumii}{\theenumi.\arabic{enumii}.}
\usepackage{etoolbox}
\makeatletter
\patchcmd{\@makecaption}
{\scshape}
{}
{}
{}
\makeatother

\title{\LARGE \bf
Leveraging Structural Information to Improve Point Line Visual-Inertial Odometry
}
\author{ Bo Xu$^{1}$, Peng Wang$^{2}$, Yijia He$^{3}$, Yu Chen$^{1}$, Yongnan Chen$^{2}$, Ming Zhou$^{2}$
\thanks{$^{1}$Bo Xu, Yu Chen are with School of Geodesy and Geomatics, Wuhan University, Wuhan 430079, China; Corresponding author: Bo Xu, Email: boxu1995@whu.edu.cn }
\thanks{
$^{2}$Peng Wang, Yongnan Chen and Ming Zhou are with Faculty of Robot Science and Engineering, Northeastern University, Shenyang 110819, China}
\thanks{
$^{3}$Yijia He is with Kuaishou Technology, Beijing 100000, China}
}

\begin{document}

	\twocolumn[
	
		\begin{center}
		This paper has been accepted for publication in \emph{IEEE Robotics and Automation Letters}. \vspace*{13pt}
		
    	
    	\vspace*{23pt}
		\end{center}
	
      \hspace{0.5em}\copyright 2022 IEEE. Personal use of this material is permitted. Permission from IEEE must be  obtained for all other uses, in any  current or future media, including reprinting/republishing this material for advertising or promotional purposes, creating new collective works, for resale or redistribution to servers or lists, or reuse of any copyrighted component of this work in other works.
]
	\clearpage

    \maketitle
	\thispagestyle{empty}
	\pagestyle{empty}

\begin{abstract}
Leveraging line features to improve the accuracy of the SLAM system has been studied in many works. However, making full use of the characteristics of different line features (parallel, non-parallel) to improve the SLAM system is rarely mentioned.
In this paper, we designed a VIO system based on points and straight lines, which divides straight lines into structural straight lines (that is, straight lines parallel to each other) and non-structural straight lines. In addition, in order to optimize the line features effectively, we only used two parameters to minimize the representation of the structural straight line and the non-structural straight line. Furthermore, we designed a straight line matching strategy based on 2D-2D and 2D-3D matching methods to improve the success rate of straight line matching. We conducted ablation experiments on synthetic data and public datasets, as well as compared our algorithm with state-of-the-art algorithms. The experiments verified the combination of different line features can improve the accuracy of the VIO system, and also demonstrated the effectiveness of our system.
\end{abstract}

\section{INTRODUCTION}
\label{sec:introduction}
Simultaneous motion estimating and mapping is widely used in the field of intelligent robots, such as autonomous driving, rescue, and augmented reality. With camera and inertial measurement unit (IMU) being low-cost and efficient sensors, the visual-inertial odometry system (VIO) can overcome the shortcomings of the two sensors and improve the accuracy and robustness of localization. The existing VIO systems mainly use points as visual features to estimate ego-pose and build a sparse map of 3D points\cite{ORBSLAM3,VINS,Leutenegger2014}. However, in some textureless or illumination challenging environments, point-based VIOs may fail in pose estimation \cite{goodfeature}.

Line features exist widespreadly in man-made environments, which can provide additional visual constraints and build a map with richer information for automatic navigation. Thus, VIO systems based on point and line features have attracted widespread attention. Current line features can be divided into two categories based on the type of line features: non-structural lines \cite{Smith,Pumarola2017,PL-VIO,xu2020improved } and structural lines\cite{kottas2013exploiting,structslam,Zhang2012,Lee2009}. The non-structural lines have more universality and robustness because they can be operated in different kinds of environments,  but non-structural lines have no effective directional constraints as structural lines do. Structural lines can be  found in man-made environments, which are abstracted as a set of blocks sharing three common dominant directions, known as Manhattan world \cite{coughlan1999manhattan}, they encode the global orientations of the local environments, these constraints can reduce the drift of the yaw angle, however, when the environment is complex, the constraints are not effective for pose estimation.
 

\begin{figure}[t]
	\centering
	\includegraphics[width=0.9\linewidth]{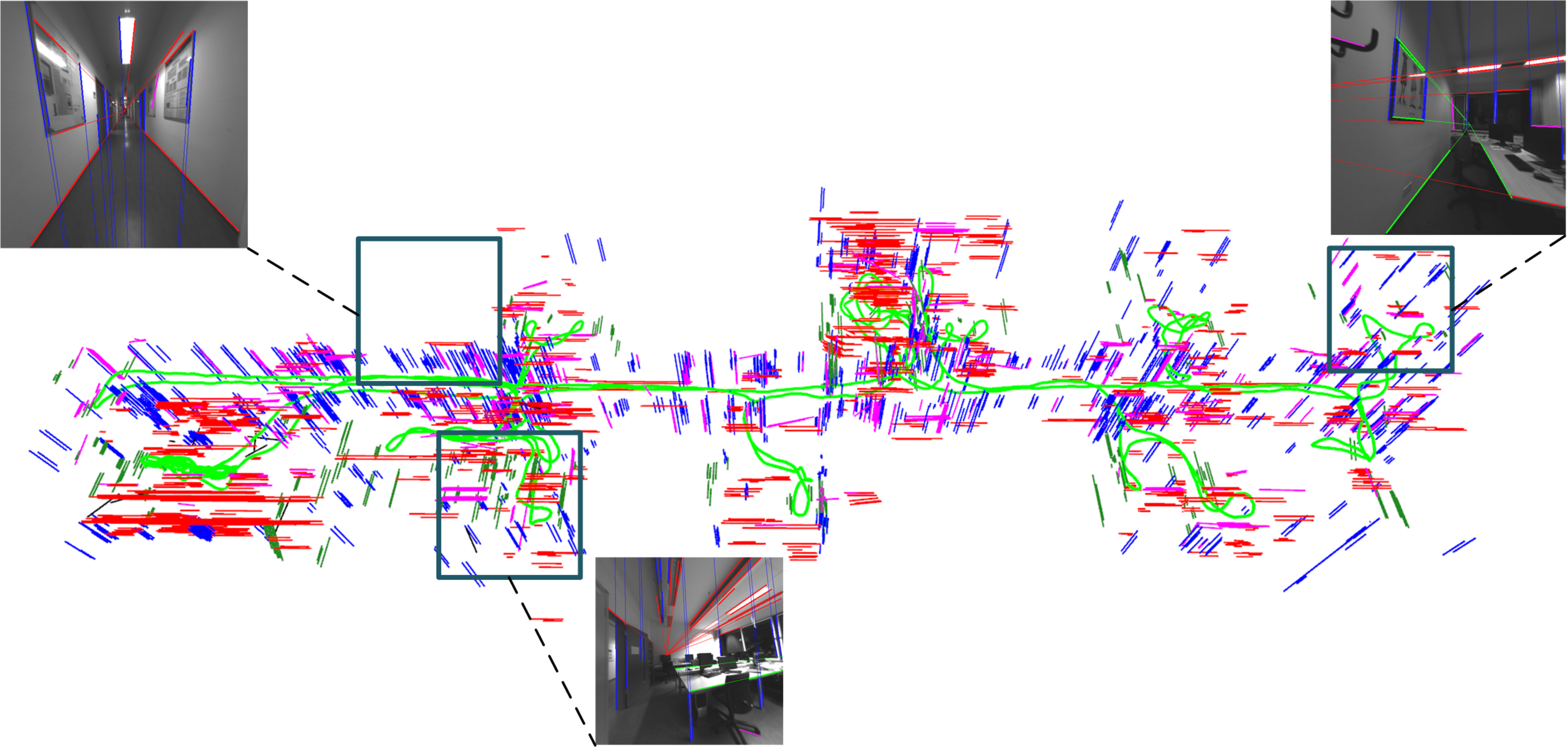}
	\caption{The proposed monocular VIO system that builds line map. Red, green and blue lines in the map are the landmarks of structrual line feature in X, Y, Z direction, the purple lines are the landmarks of non-structural line feature. Three sub-images show the lines detected in the image.}
	\label{fig:cover_grap}
	\vspace{-6mm}
\end{figure}

In order to overcome the shortcomings of insufficient visual information in the point-based VIO system, the line features are a good supplement. 
PL-VIO proposed by He et al. \cite{PL-VIO} integrates straight lines into the VIO system based on point features. In this work, the straight lines are represented with Plücker coordinates, and a minimal four-parameter orthonormal representation is used for optimization because the Plücker coordinates are over-parameterized. The Trifo-VIO system proposed by Zheng et al. \cite{Tsai2018} uses Kalman filter to integrate the straight lines into the VIO system. The authors use geometric constraints expressed by the normal vector of line to construct the constraints. 

{However, none of them distinguish the difference between structural and non-structural lines, and the parallel constraints between structural lines are not adopted.} Camposeco et al. \cite{camposeco2015using}  incorporated  vanishing points into the VIO system, using global constraint information of vanishing points to correct the yaw angle drift in pose estimation. However, the system only regards structural lines as intermediate results and does not make full use of  structural lines to correct translation. Zou et al. \cite{Zou2019} proposed a new parameter form of structural lines and integrated the structural line features into the VIO system, but non-structural line features are not employed.

Overall,  non-structural lines and structural lines have the potential to improve the performance of the VIO system with different advantages, by leveraging both of them, the estimator will be more robust and accurate in the complex environment. This paper proposes a tightly coupled monocular VIO system named PLS-VIO (Point, Non-structural line, Structural line VIO), including visual point features, non-structural line features, and structural line features to achieve accurate pose estimation and point-line map construction, as shown in Fig.\ref{fig:cover_grap}. To reduce the computational burden  in the process of line optimization, we introduce 2-parameter expression for non-structural line features. Furthermore, we describe the classification, matching, and initialization of all the lines in detail. The main contributions of this work include:

\begin{itemize}
	\item We divide lines observed by the VIO system into structural lines and non-structural lines, compared with the single type of line features, our method utilizes the different line constraints to improve the accuracy and robustness of pose estimation and mapping in the complex environment. 
	
	
	\item We design a 2D-2D and 2D-3D line matching algorithm to reduce mismatching and tracking lost in a long period of line features.  In order to optimize the line features efficiently, we introduce a 2-parameter representation of non-structural lines, the line features are fused efficiently into the optimization-based estimator.
	
	\item We provide  ablation experiments to verify the effectiveness of the proposed algorithm on the synthetic data,  EuRoc dataset \cite{Burri2016}, and TUM VI benchmark \cite{Schubert2018}. Experimental results demonstrate that our system is capable of accurate  pose estimation and mapping.
\end{itemize}

\section{SYSTEM OVERVIEW}\label{sec:system overview}

	The system proposed in this paper is based on VINS-Mono\cite{VINS}. VINS-Mono uses the optimization method to tightly couple the IMU observation and the visual observation of point features.  Our system adds non-structural lines and structural lines, as well as constructs corresponding constraints. 
	
\begin{figure}[htpb]
	    \centering
	    \includegraphics[width=.7\linewidth]{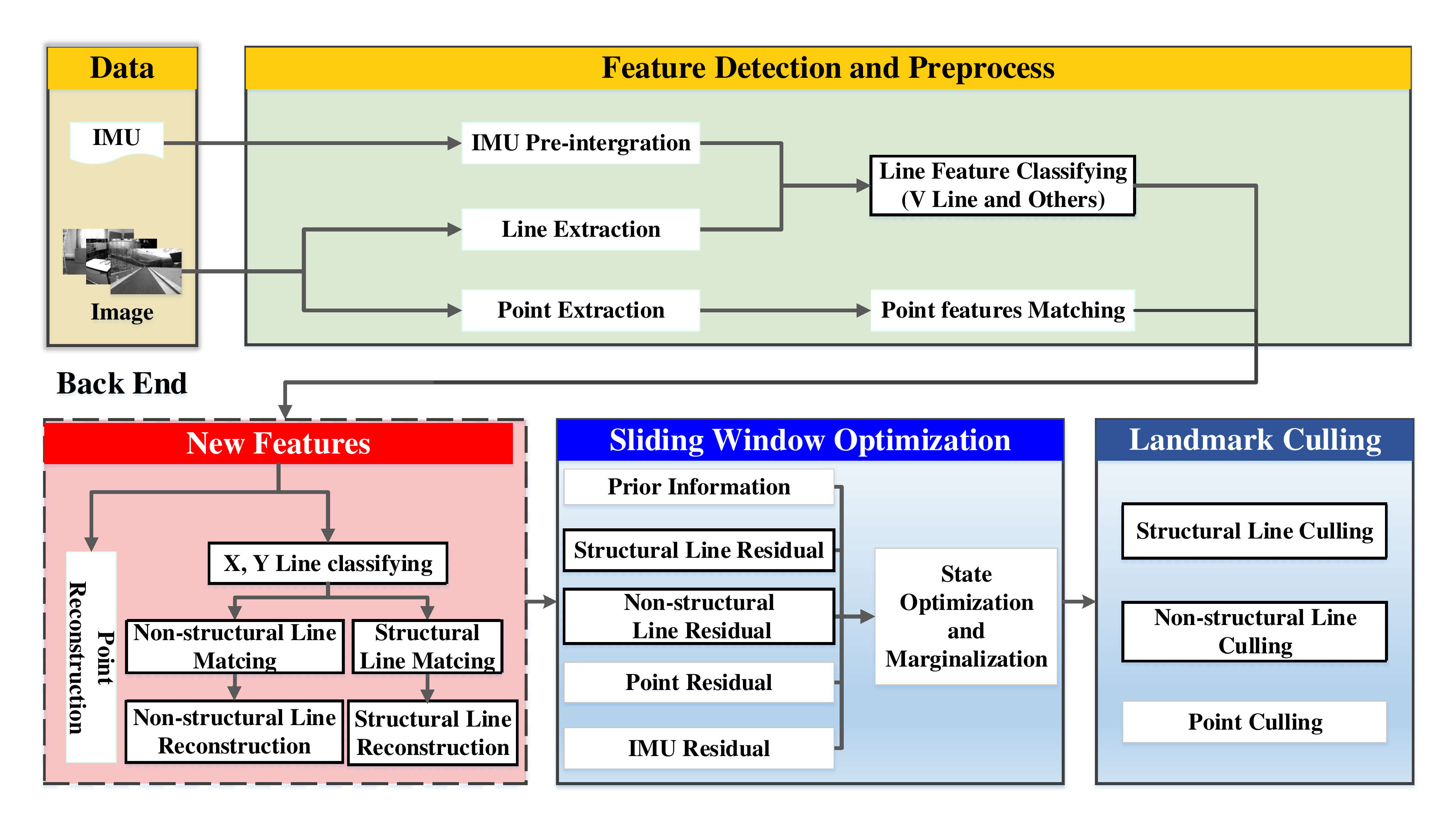}
	    \caption{Overview of our PLS-VIO system. The Front-End module is used to extract information from the raw measurement; Line classification and state variables are estimated with sliding window optimization in the Back-End.}
	    \label{fig:pipeline}
	    \vspace{-3mm}
	\end{figure}

	As shown in Fig.\ref{fig:pipeline}, our system contains two modules: the front end and the back end. In the front end, raw measurements of IMU and image are pre-processed, including IMU pre-integration, point detection and matching, line detection and vertical line classification.
	
	In the back end, the operations for the non-structural lines and structural lines are mainly introduced. We pass the vertical line which is aligned with gravity direction and non-vertical lines to the back end, X, Y direction and non-structural lines from non-vertical direction lines are further classified, this will be described in \ref{alg: structural line and non-structural line classification}. Nextly, two different line matching strategies are operated, for the efficiency of the code operation and the simplification of the line data management, we move the matching of lines to the back end, this process will be described in \ref{alg: Line matching for 2D-2D and 2D-3D}. After this, we initialize the lines to get 3D line landmarks, which will be introduced in the \ref{alg: Line Initialization for Structural Line and Non-structure Line}. Finally, the IMU body state and 3D landmarks in the map will be optimized by minimizing the sum of the IMU residual, prior residual, point re-projection residual, and line re-projection residual, all these residuals will be introduced in the \ref{sec: VIO with Lines and Points}.

	\textbf{}

	\section{Structural and Non-structural Line Methodology} \label{sec:algorithm}
	
%
  
In this section,  we introduce the implementation details of non-structural lines and structural lines in the VIO system. First, the parameter representation of point and line landmarks is described. Second, we introduce the classification of the lines. Finally, the different matching methods of lines and the initialization for non-structural lines and structural lines are presented.

\subsection{Landmarks Representation} \label{alg: landmark representation}

\subsubsection{Point representation}
\label{alg: feacture track}
We use the inverse depth  $ \mathcal{\lambda} \in \mathbb{R}$ to parameterize the point landmark from the first keyframe in which it is observed. Given the point observation $\textbf{z} =\begin{bmatrix} u, v, 1 \end{bmatrix}^{\rm T}$ in the normalized image plane, the 3D position of landmark is obtained by $ \textbf{f} = \frac{1}{\lambda} \cdot {\textbf{z}}$ .

\subsubsection{Non-structural line representation}
\begin{figure}[ht]
	\centering
	\includegraphics[width=0.35\linewidth]{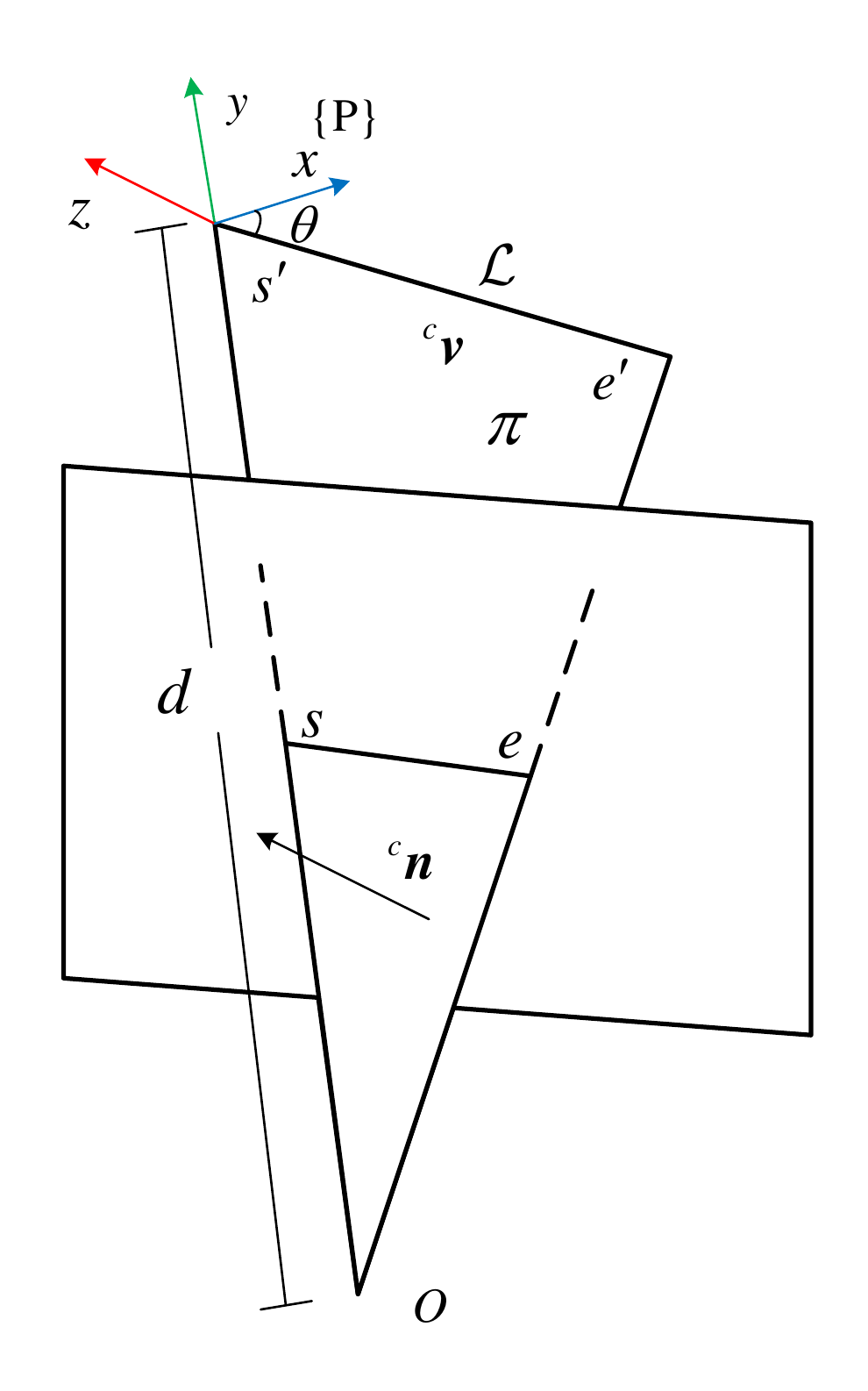}
	\caption{the endpoint ${\textbf{s}}'$ and $\textbf{e}'$ of 3D line $\mathcal{L}$ are projected to the image plane to form $\textbf{s}$ and $\textbf{e}$. $\textbf{s}'$, ${\textbf{e}}'$ and optical center ${O}$ of camera frame construct the plane $\pi$, the direction vector of $\mathcal{L}$ is ${^c}{\mathbf{v}}$, the normal vector of $\pi$ is ${^c}{\textbf{n}}$, the distance from $O$ to ${\textbf{s}}'$ is $d$, we define local coordinate system $\{P\}$ of which the origin is ${\textbf{s}}'$, the angle between the ${^c}{\textbf{v}}$ expressed in the $\{P\}$ and x axis of  the  $\{P\}$ is $\theta$.}
	\label{fig:non_structural_line}
	\vspace{-5mm}
\end{figure}

For the non-structural line, as shown in Fig. \ref{fig:non_structural_line}, the plane $\pi$ is composed of the two endpoints $\textbf{s}'\in\mathbb{R}^3$ and  $\textbf{e}'\in\mathbb{R}^3$ of the 3D line $\mathcal{L}$ and the optical center $O$ of the camera, the 3D line $\mathcal{L}$  can be expressed by $\mathcal{L} =\begin{bmatrix} {{^c}\textbf{n}}^{\rm T}, {{^c}\textbf{v}}^{\rm T} \end{bmatrix}^{\rm T}$, where ${{^c}\textbf{n}} \in \mathbb{R}^3$ is the normal vector of $\pi$, ${{^c}\textbf{v}} \in \mathbb{R}^3$ is the direction vector of $\mathcal{L}$. Due to the noise of the host frame that line is anchored is small, we fix the normal vector ${{^c}\textbf{n}}$ of the plane where the line lies,  and use the parameters of 2 DOF to represent the 4 DOF line. The local coordinate system on the plane $\pi$ is defined as  $\{P\}$, to simplify the representation of $\mathcal{L}$, we let the origin of $\{P\}$ be $\textbf{s}'$, then let the direction of y axis be aligned to the ray passing from $O$ to $\textbf{s}'$ and let direction of z axis be parallel to  the normal vector ${^c}\textbf{n}$ of plane $\pi$. Due to the orthogonality of the coordinate axes, the x axis is perpendicular to the y axis and the z axis. The distance from $O$ to $\textbf{s}'$ is $d \in \mathbb{R}$. 

To reduce the number of line parameters during optimization, we propose to use a compact parameterization that has only two parameters: $\theta$ and $\rho = 1/d$. The $\theta$ is the angle between line direction ${{^c}\textbf{v}}'$ in local coordinate $\{P\}$  and the x axis of the $\{P\}$. The ${{^c}\textbf{v}}'$ can be obtained by ${{^c}\textbf{v}}' = \textbf{R}_C^P {{^c}\textbf{v}}$, where $\textbf{R}_C^P \in \mathbb{R}^{3\times3}$ is the rotation matrix of camera coordinate w.r.t the local coordinate $\{P\}$.

\textbf{}	
\subsubsection{Structural line representation}
\begin{figure}[t]
	\centering
	\includegraphics[width=0.65\linewidth]{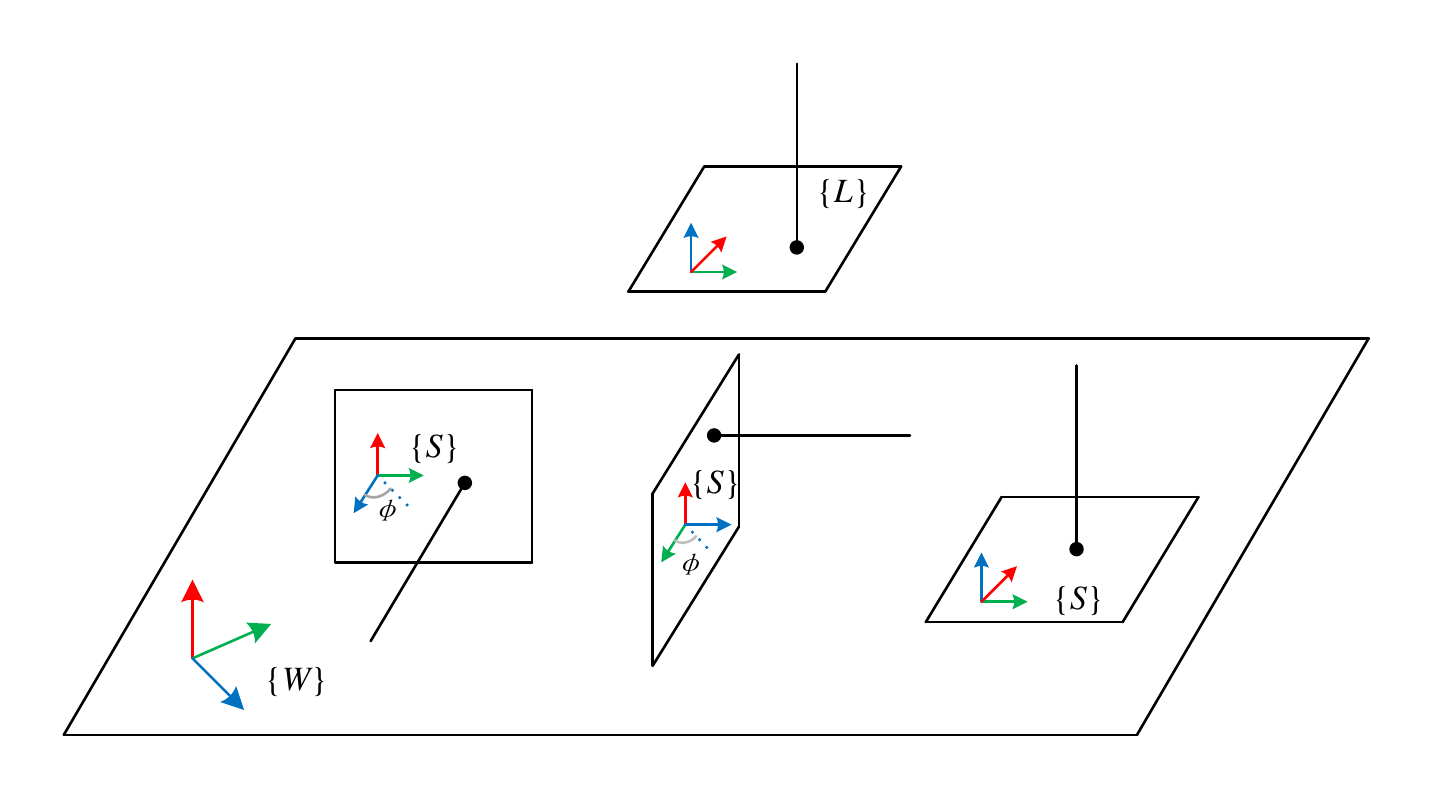}
	\caption{Start frames $\{S\}$ of X, Y, Z direction structural lines    from the parameter space $\{L\}$, the origin of start frame is in the world coordinate system $\{W\}$, the orientation of start frame is rotated about $\phi$ from the world coordinate system. }
	\label{fig:strcutural_line}
	\vspace{-5mm}
\end{figure}

We represent structural lines in multiple local Manhattan worlds with different orientations \cite{Zou2019}. Specifically, as shown in Fig. \ref{fig:strcutural_line}, each structural line is anchored to the local coordinate system where it is first observed, and we define this anchored coordinate system as a start frame $\{S\}$. The rotation and translation of start frame $\{S\}$ w.r.t the world coordinate is
 $( \textbf{R}_S^W, \textbf{P}_S^W )$. The $\textbf{R}_S^W$ is the rotation matrix of associated local Manhattan world frame w.r.t world frame, which is rotated about $\phi$ from the world coordinate system.  The $\textbf{P}_S^W$ is same as the position of camera coordinate w.r.t world frame in which the line is firstly observed.

 In order to express uniformly the X, Y, Z direction lines in $\{S\}$, we define again a parameter space $\{L\}$, the transformation from the parameter space $\{L\}$ to the start frame $\{S\}$ is a pure rotation $\textbf{R}_L^S$. In the parameter space, the structural line can be represented as the intersection point $\textbf{l}_p^l =\begin{bmatrix} a, b, 0 \end{bmatrix}^{\rm T} $ on the XY plane, we use the inverse depth to represent the intersection point, namely $\begin{bmatrix} \theta, \rho, 0 \end{bmatrix}^{\rm T}$ , where $\rho = \sqrt{a^2 + b ^2}$ and $\theta = atan2(b, a)$, we adopt this representation to speed up the convergence of structural line.

\subsection{Structural Line and Non-structural Line Classification} \label{alg: structural line and non-structural line classification}


We use the LSD algorithm\cite{Morel2010} to detect lines in the image and then classify these straight lines into structural lines and non-structural lines. The vanishing points  \cite{camposeco2015using} in the image are used to recognize the structural lines.  We detect the vanishing point of Z direction with the aid of IMU information \cite{Zou2019}, for structural lines of Z direction, we draw a ray  from the vanishing point $\textbf{v}_z$ of Z direction to the middle point of line segment $S$. The angle $A_{err}$ and distance $D_{err}$ between the ray and $S$ are further calculated. If $A_{err}$ is less than the angle threshold $A_{th}$, $D_{err}$ is less than the distance threshold $D_{th}$, we consider $S$ to be the structural line of Z direction. For the remaining lines, we first attach the line to the detected Manhattan world according to the consistency of angle and distance, when the number of unclassified lines is greater than $60\%$ of the total number of detected lines, we use the RANSAC method \cite{camposeco2015using} to detect the vanishing points in X and Y directions, and then classify lines into the new Manhattan world similar to the method of Z direction line classification, finally, the lines that are not classified are considered as non-structural lines.

\label{key}

\subsection{Line Matching for 2D-2D and 2D-3D } \label{alg: Line matching for 2D-2D and 2D-3D}
 To improve the stability and accuracy of line tracking, we combine two tracking strategies, namely frame-to-frame line tracking and frame-to-map line tracking. In general, we perform frame-to-frame tracking  method to track new detected lines, if the number of matched lines is too small, we will perform the frame-to-map method to increase the number of the matched lines. Compared with the matching method in \cite{Zou2019}, the main aim of performing firstly frame-to-frame matching method is to avoid mismatching due to inaccurate position estimation of 3D line landmarks.


\subsubsection{Frame-to-frame line tracking method}
For the frame-to-frame line tracking method, we sample all lines in the previous frame and get the set of sampling points $p_i \in \{P \_ {sample}\}$, and then use the epipolar searching method\cite{Engel2017} to find corresponding candidate matching points in the current frame. The ZMSSD (Zero-mean Sum of Squared Differences) template \cite{vincke2012real} is used to calculate the matching score of two points. We select the candidate point with the highest matching score as the tracked point and get the set of tracked points  $p'_i \in \{P \_ {tracked}\}$. The tracked point is valid if the distance from it to the line of the current frame is less than the threshold $m\_th$, $m\_th$ is set to be 5 pixels in our implementation. We consider the line to be the best matching if the number of valid points is greater than 0.8 times of sampling points on the line of the previous frame.

\subsubsection{Frame-to-map line tracking method}
For the frame-to-map line tracking, we use the ZNCC (Zero-normalized cross-correlation) matching method \cite{Stefano2005}. For a 3D line, we get the latest observation frame ${F}_i$ in the history frames corresponding to the line,  the lines between ${F}_i$ and the current frame are matched using ZNCC method.  Due to the rapid movement of the camera and the occlusion in the scene, the lengths of lines to be matched are quite different, which affects the accuracy of the matching. We use epipolar geometric constraints to determine the sampling range of the line to assist ZNCC matching, that is, after calculating essential matrix $\textbf{E}$ between ${F}_i$ and the current frame, the line's endpoints in ${F}_i$ are projected to the current frame according to the epipolar geometry, the  corresponding sampling range is determined by intersecting two epipolar lines with matching line in the current frame,which improves the success rate of ZNCC matching method.



\subsection{Initialization for Structural Line and Non-structure Line } \label{alg: Line Initialization for Structural Line and Non-structure Line}
The stability and accuracy of the line initialization have great effects on the pose estimation. For the two parameterized expressions of structural lines and non-structural lines, we use different initialization methods to determine reasonable initial values.
\subsubsection{Initialization of non-structural line}
The line segment in the normalized image plane can be represented by two endpoints $\textbf{s}^{c_1}=\begin{bmatrix} u_s, v_s, 1 \end{bmatrix}^{\rm T}$ and $\textbf{e}^{c_1}=\begin{bmatrix} u_e, v_e, 1 \end{bmatrix}^{\rm T}$. Three non-collinear points, including two endpoints of a line segment and the optical center $O$ of camera, determine a plane $ \bm{\pi} =\begin{bmatrix} \pi_x, \pi_y, \pi_z, \pi_w\end{bmatrix}^{\rm T}$, given the two plane $\bm{\pi_1}$ and $\bm{\pi_2}$ in the camera frame $c_1$, the dual Plücker matrix ${^{c}}\mathcal{L}^{*}$ can be computed by:
 \begin{flalign} 
 	{^c}\mathcal{L}^{*}=\begin{bmatrix} [^c\textbf{v}]_{\times} & ^c \textbf{n} \\ -{^c\bm{n}}^{\rm T}&0 \end{bmatrix} = \bm{\pi}_1 \bm{\pi}_2^{\rm T} -  \bm{\pi}_2 \bm{\pi}_1^{\rm T}  \in \mathbb{R}^{4 \times 4}
 \end{flalign}
where $[\cdot]_{\times}$ is the skew-symmetric matrix of a three-dimensional vector.

We get the Plücker coordinate of the line ${^c}\mathcal{L} = \begin{bmatrix} {^c}\textbf{n}^{\rm T}, {^c}\textbf{v}^{\rm T} \end{bmatrix}^{\rm T} $from the dual Plücker matrix, where ${^c}\textbf{n}\in \mathbb{R}^3$ denotes the normal vector of the plane determined by ${^c}\mathcal{L}$ and the origin of the camera frame $c_1$.  The ${^c}\textbf{v} \in \mathbb{R}^3$ denotes the direction vector determined by the two endpoints of ${^c}\mathcal{L}$.  We transform ${^c}\textbf{v}$ in the camera coordinate into local coordinate $\{P\}$ by ${{^c}\textbf{v}}' = R_C^P {^c}\textbf{v}$, where $R_C^P = {R_P^C}^{\rm T}$, the columns of ${R_P^C}$ is composed of the x, y, z axis of local coordinate, respectively. Therefore, we can initialize the $\theta$ as the angle between the x axis direction of the local coordinate system and ${{^c}\textbf{v}}'$, and $\rho$ is the expression of the inverse depth, and is generally initialized as $\rho_0 = 0.2$.

\subsubsection{Initialization of structural line}
The initialization of the structural line also needs to first calculate the Plücker coordinate of the line ${^c}\mathcal{L} = \begin{bmatrix} {^c}\textbf{n}^{\rm T}, {^c}\textbf{v}^{\rm T} \end{bmatrix}^{\rm T} $. And then we get the three-dimensional endpoints expression of line in the world coordinate ${^w}\mathcal{L} = \begin{bmatrix} {\textbf{s}^w}^{\rm T}, {\textbf{e}^w}^{\rm T} \end{bmatrix}^{\rm T} $ using line trimming \cite{zhang2015building}. To obtain the intersection point of ${^w}\mathcal{L}$ and XY plane in the world coordinate system, we transfer  ${^l}\textbf{p}$  to the world coordinate system to get ${^w}\textbf{p}$  through formula (\ref{transfer parameter direction}):

\begin{flalign} 
	{^w}\textbf{p} = \begin{bmatrix} \textbf{R}_S^W \textbf{R}_L^S  & \textbf{P}_S^W   \\ \textbf{0} & 1 \end{bmatrix}^{-\rm T}{^l}\textbf{p}  \label{transfer parameter direction}
\end{flalign}

Then we intersect ${^w}\mathcal{L}$ with the plane ${^w}\textbf{p}$ to get the intersection point ${^w}\textbf{l}_p$ , and transfer ${^w}\textbf{l}_p$ to the parameter space  to get point ${^l}\textbf{l}_p$ in return.
\begin{flalign} 
	{^l}\textbf{l}_p = {\textbf{R}_L^S}^{\rm T} {\textbf{R}_S^W}^{\rm T} ({^w}\textbf{l}_p - \textbf{P}_S^W)    \label{line intersection}
\end{flalign}

We use the intersection  ${^l}\textbf{l}_p =  \begin{bmatrix} {^l}l_{px}, {^l}l_{py}, 0 \end{bmatrix}^{\rm T}$ to initialize the structural line. $\theta$ can be initialized as $\theta_0 = atan2({^l}l_{py}, {^l}l_{px})$ . The inverse depth can be initialized as $\rho_0 = 1/\sqrt{{{^l}l_{px}}^2 +{{^l}l_{py}}^2 }$.
\section{VIO with Line and Point } \label{sec: VIO with Lines and Points}

In this section, we will fuse the IMU and visual information with the sliding window optimization to build VIO system which estimates body states and 3D landmarks.
\subsection{VIO System Formulation } \label{alg: VIO system formulation}

We optimize all the state variables in the sliding window by minimizing the sum of cost terms from IMU residual, visual residual and prior residual:

\begin{flalign} 
\begin{array}{c}
	\bm{\chi}=\arg \min \limits_{\bm{\chi}} \|  \mathbf{r}_{p}-\mathbf{J}_{p} \bm{\chi} \|^{2}+ 
	\sum \limits_{i \in\mathcal{B}} \rho\left(\left\|\mathbf{r}_{b}\right\|_{\sum_{b, b_{i}+1}}^{2}\right) \\+ 
	\sum\limits_{(i, k) \in \mathcal{F}} \rho\left(\left\|\mathbf{r}_{f_k}^{c_i}\right\|_{\sum_{\mathcal{F}}}^{2}\right)+\sum\limits_{(i, l) \in \mathcal{L}} \rho\left(\left\|\mathbf{r}_{L_{l}}^{c_{i}}\right\|_{\sum_{\mathcal{L}}}^{2}\right) \\ + 
	\sum\limits_{(i, s) \in \mathcal{C}} \rho\left(\left\|\mathbf{r}_{c_{s}}^{c_{i}}\right\|_{\sum_{\mathcal{C}}}^{2}\right)
\end{array}
\end{flalign}
Where $\mathbf{r}_{b}$ is the IMU measurement residual, $\mathbf{r}_{f_k}^{c_i}$ is the re-projection residual of point, $\mathbf{r}_{L_{l}}^{c_{i}}$ and $\mathbf{r}_{c_{s}}^{c_{i}}$ are the re-projection of non-structural line and structural line respectively.   $\mathbf{r}_{p}$ and $\mathbf{J}_{p}$ are the prior residual and Jacobian from marginalization operator\cite{VINS}, respectively. $\rho(\cdot)$ is the Cauchy robust function used to suppress outliers. $\sum_{(\cdot)}$ is the covariance matrix of a measurement. The covariance matrix $\sum_{b, b_{i}+1}$ of IMU is calculated by covariance propagation with IMU measurement noise, the covariance matrix of visual measurement is determined by the prior knowledge. 

\subsection{Point Feature Measurement Model } \label{alg: Point Feature Measurement Model}
For point features, we use the distance between the projected point and the observed point in the normalized image plane defined as re-projection error to represent the residual. Given the $k^{th}$ point feature measurement at frame $c_j$, $\textbf{z}_{f_k}^{c_j} = \begin{bmatrix} u_{f_k}^{c_j},  v_{f_k}^{c_j}, 1 \end{bmatrix}^{\rm T}$ , the re-projection error is defined as: 
\begin{flalign} 
\begin{array}{l}
    \mathbf{r}_{f_k}^{c_i}=
	\left[\begin{array}{l}
		\frac{x^{c_{j}}}{z^{c_{j}}}-u_{{f}_{k}}^{c_{j}} \\
		\frac{y^{c_{j}}}{z^{c_{j}}}-v_{{f}_{k}}^{c_{j}}
	\end{array}\right] 
\end{array}	
\end{flalign}
where $\begin{bmatrix} x^{c_j},  y^{c_j}, z^{c_j} \end{bmatrix}^{\rm T}$ is the projected point from the first observation frame of the feature. 
\subsection{Non-structural Line Measurement Model} \label{alg: Non-structural Line Measurement Model }

For the non-structural line, we express the measurement model by transferring the line segment parameters observed in the first keyframe to the other keyframe which observes the line. The direction vector of line in the local coordinate can be calculated by ${^d}\textbf{v} = \begin{bmatrix} cos \theta, sin \theta, 0 \end{bmatrix}^{\rm T}$, we can transfer the ${^d}\textbf{v}$ from the local coordinate to the camera coordinate system to get the line direction ${^c}\textbf{v}$.
\begin{flalign} 
{^c}\textbf{v} = \textbf{R}_P^C {^d}\textbf{v}
\end{flalign}
 Where $\textbf{R}_P^C$  is the rotation matrix from local coordinate to the camera frame coordinate, the columns of $\textbf{R}_P^C$ is composed of the x, y, z axis of local coordinate $\{P\}$, respectively.
 
The one endpoint $\textbf{s}$ of the line segment in the camera frame can be calculated by 
\begin{flalign} 
	\textbf{s} =  \frac{\textbf{y}}{\| \textbf{y} \|} \cdot d
\end{flalign}
 Where $\textbf{y}$ is the y axis of the local coordinate.
 
 To obtain the projection of a line on the normalized image plane, it requires to transfer both the endpoint  $\textbf{s}$  and the direction ${^c}\textbf{v}$ in the first observation keyframe to the target keyframe. We get the endpoint $\textbf{s}'$ and direction ${{^c}\textbf{v}}'$ in the target frame by: 
 \begin{flalign} 
 \begin{array}{l}
\textbf{s}'=
 \left[\begin{array}{l}
 	x' \\
 	y' \\
 	z'
 \end{array}\right]
= {\textbf{R}^W_{C_j}}^{\rm T}({\textbf{R}^W_{C_i}}\textbf{s} + {\textbf{P}^W_{C_i}}) -  {\textbf{R}^W_{C_j}}^{\rm T}{\textbf{P}^W_{C_j}}
\end{array}	
\end{flalign}

\begin{flalign} 
 {{^c}\textbf{v}}'= {\textbf{R}^W_{C_j}}^{\rm T}{\textbf{R}^W_{C_i}}{^c}\textbf{v}
\end{flalign}
where $(\textbf{R}^W_{C_i}, \textbf{P}^W_{C_i})$ is the pose of first keyframe that observes line feature, and $(\textbf{R}^W_{C_j}, \textbf{P}^W_{C_j})$ is the pose of target keyframe that observes line features, we absorb the transformation from IMU to the camera.

We get the line equation on the target normalized image plane by
\begin{flalign} 
\textbf{l}_l^{m_i} = [\textbf{s}']_{\times} {{^c}\textbf{v}}'
\end{flalign}

The measurement of the line segment $\textbf{z}^{m_i}_{L_l}$ on the normalized image plane consists with two endpoints $\textbf{s}_l^{m_i} = \begin{bmatrix} u_s, v_s, 1 \end{bmatrix}^{\rm T}$ and $\textbf{e}_l^{m_i} =\begin{bmatrix} u_e, v_e, 1 \end{bmatrix}^{\rm T}$, the line re-projection residual is defined as:
\begin{flalign} 
\mathbf{r}_{L_{l}}^{c_{i}} = \frac{1}{\|\textbf{z}^{m_i}_{L_l}\|}\begin{bmatrix} d(\textbf{s}_l^{m_i},\textbf{l}_l^{m_i} )\\
	d(\textbf{e}_l^{m_i},\textbf{l}_l^{m_i} )
 \end{bmatrix}
\end{flalign}
With $d(\textbf{s}, \textbf{l})$ is the distance from the endpoint $\textbf{s}$ to the projection line $\textbf{l}$:


\subsection{Structural Line Measurement Model} \label{alg: Structural Line Measurement Model}
The residual form of the structural line is to transfer the line parameter in the parameter space $\{L\}$ to the start frame $\{S\}$, and then to the other keyframe that observe the 3D line. In the target keyframe, the line re-projection residual is constructed.

Using (\ref{structural line intersecton transform}), the parameters of the line are transferred from the parameter space of the first observation keyframe to the camera coordinate of the target keyframe, in order to simplify the formula, we absorb the transformation from the IMU to the camera.
\begin{flalign} 
	 {^c}\textbf{l}_p = \textbf{R}_W^C \textbf{R}_S^W \textbf{R}_L^S {^{l}}\textbf{l}_p + (\textbf{R}_W^C \textbf{P}_S^W + \textbf{P}_W^C)  \label{structural line intersecton transform}
\end{flalign}
where $(\textbf{R}_S^W,\textbf{P}_S^W)$ represents the transformation from the start frame to the world coordinate, $(\textbf{R}_W^C,\textbf{P}_W^C)$ represents the transformation from the world coordinate to the target keyframe. 
${^{l}}\textbf{l}_p  = \begin{bmatrix} a, b, 0 \end{bmatrix}^{\rm T} $, $a, b$ are expressed by the line parameters $\rho, \theta$, with $a = \frac{cos\theta}{\rho}$, $b = \frac{sin\theta}{\rho}$.

The direction of structural line is represented in the parameter space ${^l}\textbf{v} = \begin{bmatrix} 0, 0, 1 \end{bmatrix}^{\rm T}$, and we use (\ref{structural line direction transform})  to transform it into the camera frame coordinate.
\begin{flalign} 
	{^c}\textbf{v} = \textbf{R}_W^C \textbf{R}_S^W \textbf{R}_L^S {^l}\textbf{v} \label{structural line direction transform}
\end{flalign}

We get the line equation on the target frame by:
\begin{flalign} 
	\textbf{l}_s^{m_i} = [^{c} \textbf{l}_p]_{\times} {^c}\textbf{v}
\end{flalign}

Similar to the non-structural line,  the structural line re-projection residual is also defined by the distance between the observation and the line from perspective projection.


\section{Experiments} 
\label{sec:experiment}
In order to  analyze our method, we conducted comprehensive experiments on the synthetic data, EuRoc dataset, and TUM VI benchmark dataset. Besides, we also provided
an intutive demonstration:  \textcolor{blue}{\url{https://youtu.be/OnMAsmDsTVE}}, the related code of our algorithm is released at \textcolor{blue}{\url{https://github.com/xubogithub/Structural-and-Non-structural-line}}.  In these experiments, the algorithm ran on a computer with Intel Core i7-9750H@ 2.6GHz, 16GB memory and ROS Kinetic \cite{Quigley},  ceres 2.0.0 \cite{ceres-solver}.



To assess the advantages of our proposed approach, we compared our method with  VINS-Mono without loop closure \cite{VINS}, PL-VIO\cite{PL-VIO}, PL-VINS without loop closure \cite{fu2020pl}, openVINS with monocular mode \cite{geneva2020openvins}, and BASALT \cite{usenko2019visual}. We choose the absolute pose error (APE) as the main evaluation metric which directly compares the trajectory error between the estimated pose and the groundtruth. The open-source accuracy evaluation tool evo \cite{grupp2017evo} was used to evaluate the trajectory accuracy.

\subsection{Synthetic Data}
\begin{figure}[b]
	\centering
	\includegraphics[scale=0.18]{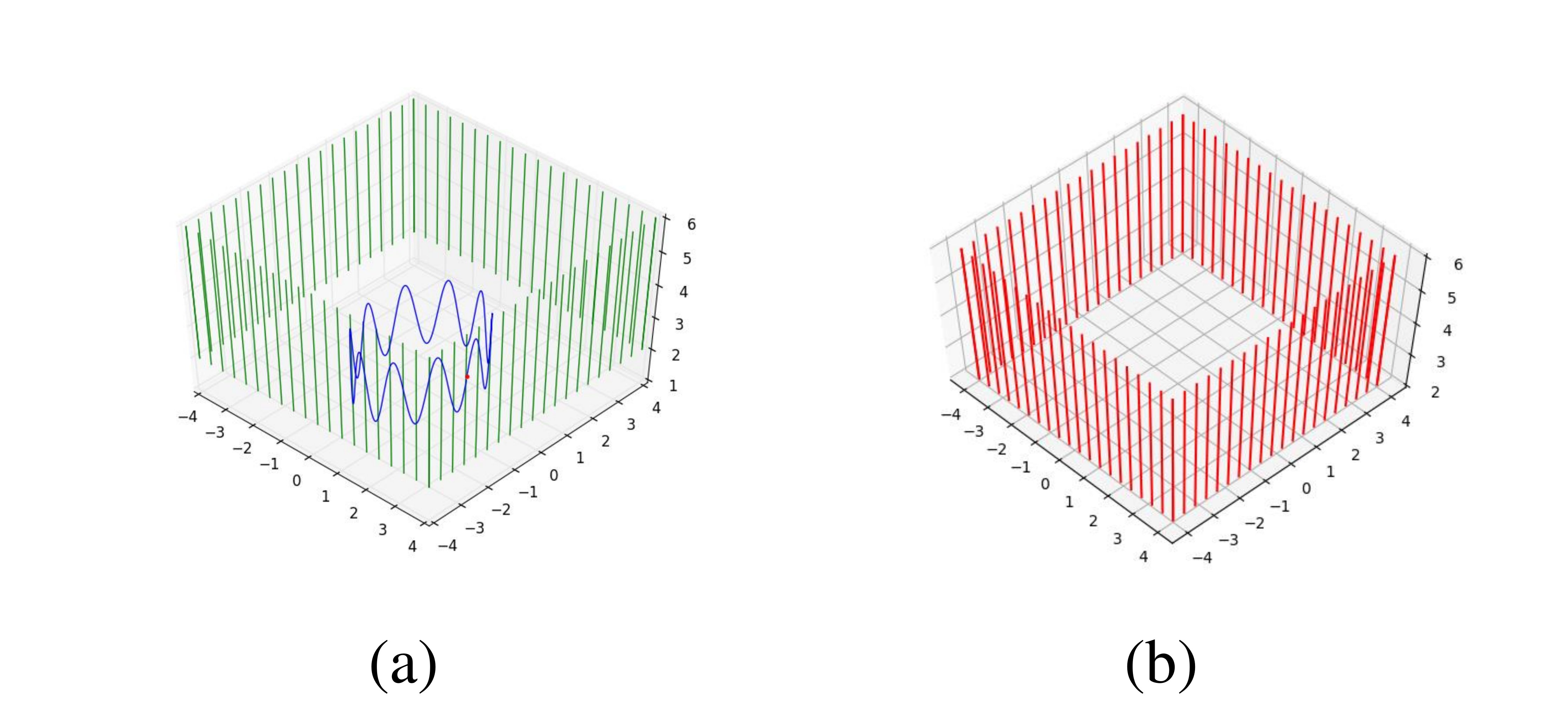}
	\caption{ Synthetic data and reconstruction of 3D lines  by our method. (a) groundtruth. (b) our method.}
	\label{fig:Synthetic line map}
	\vspace{-3mm}
\end{figure}

To verify the validity of non-structural lines' parametric expression, we generated a synthetic environment in which all 3D lines make up a room as groundtruth. As is shown in Fig.\ref{fig:Synthetic line map}, the green lines in (a) are the simulated 80 landmarks, the blue track is composed of 600 VIO poses, each pose forms line observations in the camera frame, red lines in (b) are reconstructed by our method while there is no noise added to the line observations.
\begin{table}[b]
	\label{tab:3}
	\begin{center}
		\caption{Comparison of reconstruction accuracy[cm] / time[s] of different line representation in different pixel noise level.}
		\label{tab:accuracy and time of line representation}
		\centering
		\setlength{\tabcolsep}{1.0mm}{
			\renewcommand{\arraystretch}{.9} {
				
				\begin{tabular}{ccccccccc}
					\toprule
					\multirow{2}{*}{Para} & \multicolumn{2}{c}{0 pixel} & \multicolumn{2}{c}{0.5 pixel} & \multicolumn{2}{c}{1.0 pixel} & \multicolumn{2}{c}{1.5 pixel} \\ \cmidrule(r){2-3}  \cmidrule(r){4-5} \cmidrule(r){6-7} \cmidrule(r){8-9}  
					&trans.           &time.          &    trans.        &   time.            &   trans.         &   time.            &   trans.         &    time.           \\  \midrule 
					2-para ( ours ) & 0.06 & 18.54 & 2.29 & 18.06 & 5.13 & 17.85 &  7.54 & 18.19 \\ 
					2-para (\cite{sola2009undelayed} ) & 0.06 & 18.05& 2.90 & 18.18 & 5.95 & 18.13 & 7.61 & 18.38 \\
					4-para (\cite{PL-VIO} ) & 0.05 & 36.89 & 1.80 & 36.08 & 3.65 & 36.48 & 5.41 & 35.86 \\
					\bottomrule
				\end{tabular}
		}}
	\end{center}
	\vspace{-3mm}
\end{table}
\begin{table*}[!hb]
	\label{tab:3}
	\begin{center}
		\caption{The RMSE of the different methods on EuRoc dataset. The translation (cm) and rotation (deg) error are list as follows. In \textbf{bold} the best result.}
		\label{tab:ATE Trans and Rots.}
		\centering
		\setlength{\tabcolsep}{2.0mm}{
			\renewcommand{\arraystretch}{0.6} {
				\begin{tabular}{cc|ccccccccccc}
					\toprule
					sequence&          &MH\_01  & MH\_02 & MH\_03 &MH\_04  & MH\_05 &V1\_01& V1\_02 &  V1\_03&  V2\_01& V2\_02 &V2\_03\\ \midrule
					\multirow{2}{*}{VINS-Mono}                                    & trans. &16.8 & 17.1 & 19.4 & 34.6 & 29.2 & \bf 8.7 &7.9 &20.7  & 8.2 & 15.7&20.4 \\
					& rot.  &1.4  &2.3  & 1.6 & 1.5 &\bf 0.7 & 6.3 &2.6  &6.2  &2.0  &4.3& 2.7 \\ \midrule
					\multirow{2}{*}{PL-VIO, 4-parameter} & trans. & 15.1  & 12.9 & 14.0 & 27.2 & 24.4& 9.7& 7.9 & 13.9 &8.7  & 10.4 & 13.5\\
					&rot.  &2.9 & 2.1 & 1.3 & 2.5 & 1.1 &6.1& 1.8 & 3.7 & 2.5 & 2.1 & 2.6\\ \midrule
					\multirow{2}{*}{PL-VIO, 2-parameter} &trans.  &15.5  &  13.5& 14.4 & 30.3 &  23.5& 10.5 & 8.9 &14.1  &8.3  &11.3&14.3  \\
					&rot.  &3.0  &2.5  &  1.3 &2.5  &1.3  & 6.1 & 1.8 &3.9  & 2.5 & 1.9 &2.5 \\ \midrule
					
					\multirow{2}{*}{PLS-VIO-LBD}                &trans.  & 14.0  & 10.9 & 12.5 & 14.1 & 19.4 & 10.6&7.8  & 12.8 & 9.5 & 11.4 & 12.8\\
					&rot.  &2.8  & 1.5 & \bf 1.2 & 0.9 & 1.0 &\bf 6.0  &1.5 &  3.0  & 1.0 & 2.1 & 2.8\\ \midrule        
					
					\multirow{2}{*}{PLS-VIO}                            &trans.  & \bf 12.8 & \bf 9.3  & \bf 10.6  & \bf 13.4 &  16.8 &9.8& \bf 7.3  & \bf 10.8 & \bf 7.7 & 10.6  & \bf 11.0\\
					& rot. & \bf 1.3 &  0.9  &  \bf 1.2  & 0.9 & \bf 0.7  & \bf 6.0& \bf 1.3 & 2.3 &\bf 0.6  & \bf 1.5&\bf 1.5 \\ \midrule                                                                             
					
					\multirow{2}{*}{PS-VIO} &trans.  &12.9  & 9.4 & 10.9 & 13.6& \bf 16.0& 10.3  & 8.5 &  11.9&9.7  & \bf 10.3 & 13.1 \\
					&rot.  &1.4  & \bf 0.6 & 1.4 &\bf0.8 &\bf 0.7  &6.1&1.7  & \bf 2.1 &1.0  &1.9 &1.7  \\ \bottomrule
				\end{tabular}
		}}
	\end{center}
	\vspace{-3mm}
\end{table*}
 In the experiment, we optimized all 3D line landmarks with our 2-parameter expression, the 2-parameter expression in \cite{sola2009undelayed}, and the 4-parameter expression in \cite{PL-VIO} respectively. The reconstruction accuracy is obtained by comparing the end-points of reconstructed straight lines with the simulated 3D line landmarks. As shown in Tab.\ref{tab:accuracy and time of line representation},  our 2-parameter expression has comparable accuracy compared with the 4-parameter expression, but the calculation time in half and the efficiency is greatly improved, the method in \cite{sola2009undelayed} optimizes two coefficients corresponding to the orthonormal bases of the line's direction vector, our method has the same accuracy and time consuming compared with it, which shows that the 2-parameter expression of the line can improve the efficiency of line optimization.

\subsection{EuRoc Dataset}
The EuRoc micro aerial vehicle (MAV) datasets consist of two scenes, a machine hall and an ordinary room, which contain structure and non-structure scenes, In our experiments, we used  the images from the left camera, besides, the extrinsic and intrinsic parameters are specified by the dataset.
\subsubsection{Localization Accuracy}
To illustrate the effect of the introduced structural line constraints and non-structural line constraints in the VIO system, the behavior of different parameter expressions of non-structural line features, as well as the superiority of our line matching method compared with the LBD method \cite{zhang2013efficient}, we conducted  ablation experiments to evaluate the RMSE APE of multiple systems on the EuRoc datasets. Methods used for comparison include: VINS-Mono \cite{VINS} is only with point features, PL-VIO \cite{PL-VIO} combines points and non-structural line features, where LBD descriptors are used to match lines, the 4-parameter \cite{PL-VIO} and our 2-parameter representation of straight line are implemented in PL-VIO to compare the performance of different parameters, PLS-VIO-LBD has introduced structural constraints into the PL-VIO with 2-parameter representation of line, PLS-VIO replaced the LBD descriptor in the PLS-VIO-LBD system with our line matching algorithm, and PS-VIO removed the non-structural line features in the PLS-VIO.
 \begin{figure}[!htb]
 	\centering
 	\setlength{\abovecaptionskip}{-1.mm}
 	\includegraphics[scale=0.09]{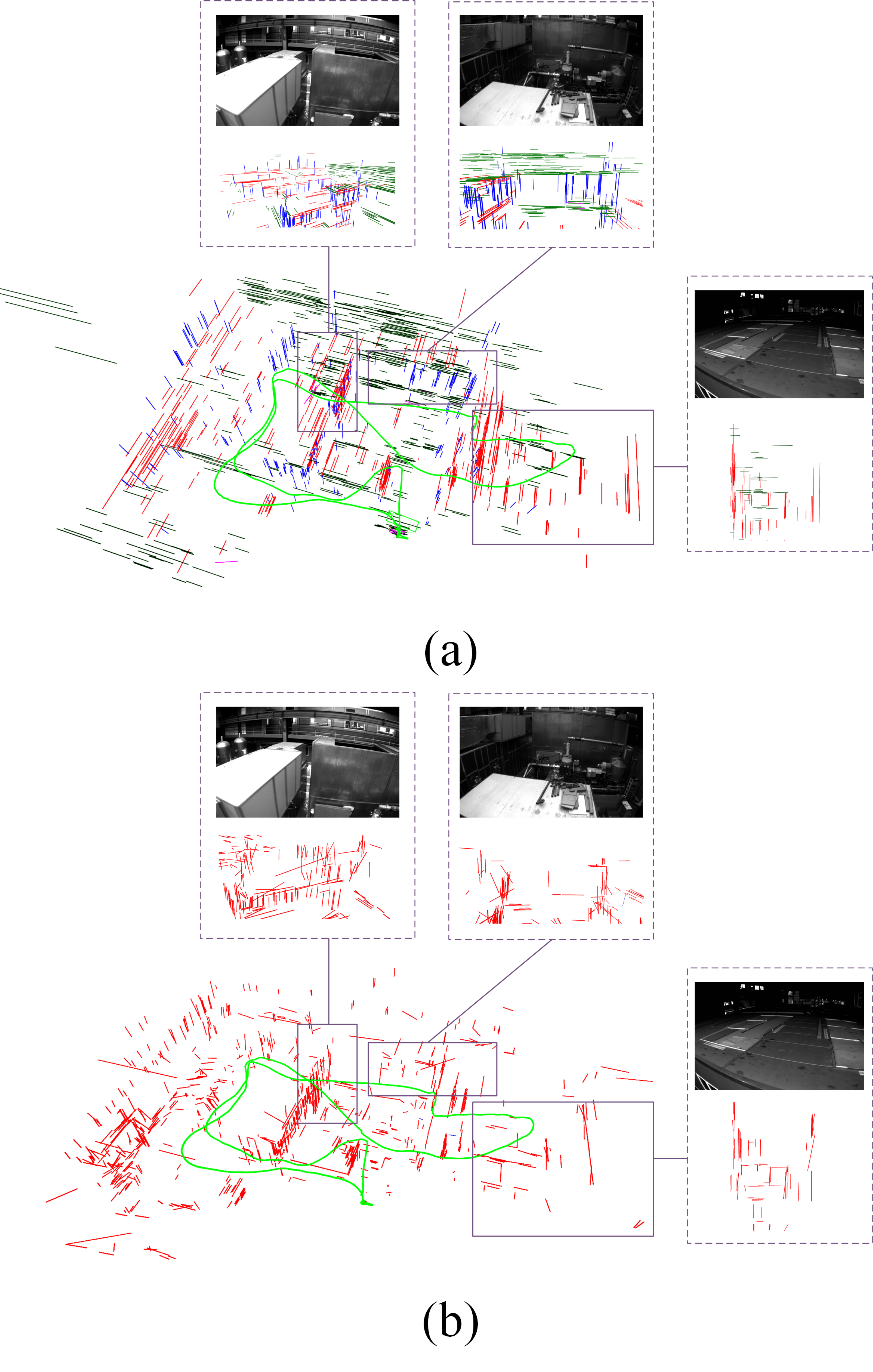}
 	\caption{Line map of EuRoc MH\_05 dataset generated by (a) PLS-VIO, red, green and blue lines in the map are landmarks of structrual line features in X, Y, Z direction, the purple lines are the landmarks of non-structural line features, (b)  PL-VIO, red lines in the map are landmarks of line features.}
 	\label{fig:PLS-PL-VIO_MAP_mh05}
 	\vspace{-3mm}
 \end{figure}
As shown in Tab.\ref{tab:ATE Trans and Rots.}, PLS-VIO achieves the smallest translation error on almost all the sequences, which means using both structural lines and non-structural lines in the VIO system can improve the accuracy and robustness of pose estimation, meanwhile, it also verifies the effectiveness of our line matching method. Specially, by comparing the experimental results of PLS-VIO-LBD with PL-VIO-2-parameter, we can find that introducing structural constraints into PL-VIO can significantly improve the accuracy of the system when there are enough structural straight lines in the scene, such as MH\_04 and MH\_05 sequences. At the same time, from the experiment results of PLS-VIO and PS-VIO, we can see that using both structural lines and non-structural lines achieves better results than only using structural lines. PLS-VIO has achieved a lower RMSE, especially on the V series, the error was reduced by an average of $11\%$. The environment of V sequences is messy and there are many non-structural lines, the constraints of which can play a better role.

\begin{figure*}[!ht]
	\centering
	\setlength{\abovecaptionskip}{-3.mm}
	\includegraphics[scale=0.13]{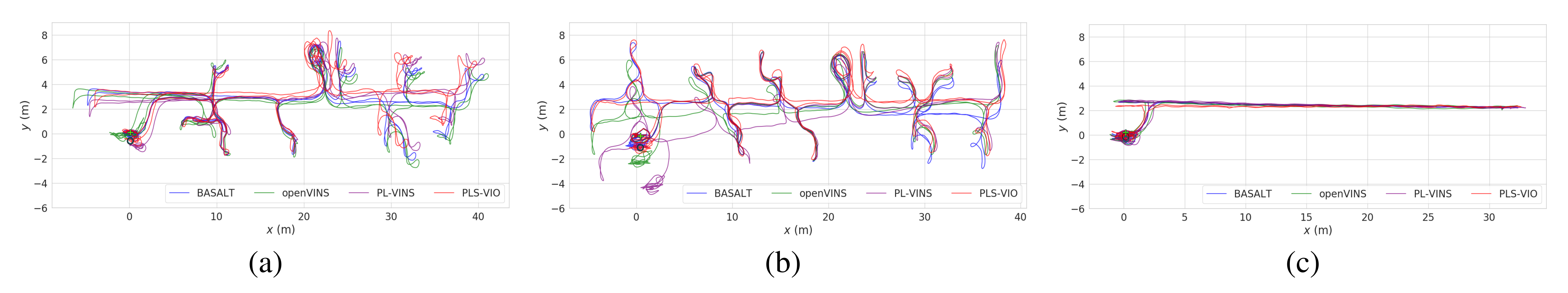}
	\begin{center}
		\caption{The trajectories of BASALT, openVINS, PL-VINS and PLS-VIO running on (a) corridor1 dataset, (b) corridor2 dataset, (c) corridor4 dataset. Cross is the start of trajectory, and circle is the end of trajectory. The corridor is along the horizontal direction, we aligned the four trajectories at the initial time with the groundtruth,  the end position of PLS-VIO is close to the groundtruth and  the drift of orientation is small.}
		\label{fig:trajectory_PLS-VIO_tum1}
		\vspace{-6mm}
	\end{center}
\end{figure*}
	Meanwhile, the expression of 2-paramater of lines is easily affected by the observation noise of straight line, the positioning accuracy of PL-VIO with 2-parameter is lower than that of 4-parameter which sacrifices accuracy to maintain efficiency. By comparing PLS-VIO and PLS-VIO-LBD, our matching method can find accurate matching lines in complex environments, thereby improving the estimation of pose.

\subsubsection{Mapping Quality}
We compared the line map constructed by PLS-VIO proposed in this paper with that constructed by PL-VIO to evaluate the map quality. Fig.\ref{fig:PLS-PL-VIO_MAP_mh05} (a) shows a bird eye view of the map for PLS-VIO, as well as three detailed sub-maps and their corresponding real scene images. The line landmarks in the map can well reflect the line features of the man-made machine hall. In order to better reflect the map quality of our system, we also run PL-VIO to build the line map, as shown in Fig.\ref{fig:PLS-PL-VIO_MAP_mh05} (b), we can observe that the map of PLS-VIO has more line features and the position of lines is also more accurate, at the same time, the accuracy improvement of the local map constructed by the VIO system will also affect the estimated pose, thereby improve the accuracy of the pose estimation.
\subsection{TUM VI Benchmark dataset}
We tested the algorithm on corridor sequences of the TUM VI benchmark dataset, the corridor scene has a typical Manhattan man-made structure and contains obvious illumination changes and weak texture, which is great challenge to the algorithms.
\begin{table}[b]
	\label{tab:4}
	\begin{center}
		\caption{The RMSE of the state-of-art methods compare to our PLS-VIO on TUM VI benchmark dataset. The translation (cm) errors are list as follows. In \textbf{bold} the best result.}
		\label{tab:ATE Trans.}
		\centering
		\setlength{\tabcolsep}{1.mm}{
			\begin{tabular}{c|cccc|c}
				\toprule
				\multicolumn{1}{c|}{Seq.}&BASALT&openVINS&	PL-VINS  & PLS-VIO& Length (m)\\ \midrule
				corridor1 &     23.2    &  63.2    &  32.5 &  \bf 18.6 & 305    \\
				corridor2 &    45.8     &   46.4   &  70.0 &  \bf 33.7 & 322\\
				corridor3 &    39.1     &    35.5  &  67.8 &  \bf 26.1 &300           \\
				corridor4 &    22.1     &    33.5  &  24.1 & \bf 21.4  & 114 \\ 
				corridor5 &    44.9     &    \bf 32.1  &  49.4 &  37.6       &270  \\ \bottomrule
				
			\end{tabular}
		}
	\end{center}
	\vspace{-2mm}
\end{table}

We ran BASALT, openVINS, PL-VINS, and PLS-VIO on corridor sequences, and drew the trajectories of four systems, openVINS and BASALT only use the point features and track points with optical flow method, which are affected by the illumination changes and weak texture environment, 
PL-VINS leverages both of the point and non-structural line features, however, due to the lack of global constraints of Manhattan regularity, the drift of
orientation is large, as is shown in Fig.\ref{fig:trajectory_PLS-VIO_tum1}.  From Tab. \ref{tab:ATE Trans.}, we can observe that our system achieves the smallest translation error on almost all the sequences, meanwhile, in the map of PLS-VIO, the landmarks of structural line feature in X, Y, Z direction are parallel to the main direction of the Manhattan world, which is shown in Fig.\ref{fig:cover_grap}, the experimental results show that our method can effectively improve the accuracy of pose estimation and mapping quality. 

\subsection{Runtime Evaluation}
\begin{table}[t]
	\label{tab:3}
	\begin{center}
		\caption{Mean execution time (Unit: millisecond) of PL-VIO with 2-parameter, with 4-parameter, PS-VIO and PLS-VIO running on the  V2\_02 sequence.}
		\label{tab:running time}
		\centering
		\setlength{\tabcolsep}{0.6mm}{
			\renewcommand{\arraystretch}{1.1}
			\begin{tabular}{ccccc}
				\toprule
				\multicolumn{1}{c}{Module}&\begin{tabular}[c]{@{}c@{}}PL-VIO\\ 2-parameter\end{tabular}&\begin{tabular}[c]{@{}c@{}}PL-VIO\\ 4-parameter\end{tabular} &	\multicolumn{1}{c}{PS-VIO} & 	\multicolumn{1}{c}{PLS-VIO}\\ \midrule
				\texttt{Line Extract  } &      15.35                 &    15.29                  &  15.21      &     15.32         \\
				\texttt{Line Match }    &       6.66                &    6.52                   &  6.49   &    6.79            \\
				\texttt{Line Classify }    &       0.00                 &    0.00                    &  0.08   &  0.08              \\ \midrule
				\texttt{Optimization}      &    11.78            &    12.85                      &    12.05      & 12.69             \\
				\texttt{Marginalization}&     7.08              &     8.25                  &   9.51        & 10.33           \\ \bottomrule
			\end{tabular}
		}
	\end{center}
	\vspace{-6mm}
\end{table}
We added structural lines and non-structural lines into the VIO based on point features, the time consuming is increased  when we extract, match and classify lines as well as optimize them in the state estimator, we compared the runtime of PL-VIO with 2-parameter line, PL-VIO with 4-parameter line, PS-VIO, and PLS-VIO, here we mainly compared the average execution time of the frontend module about line features and backend module, which are evaluated on the V2\_02 sequence. From Tab.\ref{tab:running time}, we can observe that compared with the LBD matching method, we used the 2D-2D and 2D-3D matching methods to improve the matching accuracy, but the time consumption is basically the same as that of LBD. For PL-VIO with 2-parameter and 4-parameter, the 2-parameter expression of line in this paper adds fewer state variables into the optimizer, therefore, the optimization and marginalization take less time. We added the non-structural lines into PS-VIO,  which leads to the increase of matching time, optimization time and marginalization time, we counted the total runtime of PLS-VIO, which runs at 26.12 ms in the frontend, 33.09 ms in the backend, the system can fulfill the needs of running in real-time.

\section{Conclusion} 
\label{sec:conclusion}
In this paper, we present a novel VIO system fully exploiting the point, non-structural line and structural line features, which is called PLS-VIO. Both structural lines and non-structural lines are used to improve the accuracy and robustness of mapping and pose estimation, which is superior to the  single type of lines. The 2-parameter representation of lines can speed up the optimization of the line landmarks, meanwhile, the 2D-2D and 2D-3D line matching methods  reduce the mismatching of lines. The proposed state estimator is tested in large-scale scenes and corridor environments. The experiments show that the trajectory accuracy and mapping quality of our approach are better than the state-of-the-art visual-inertial odometry.  

In the future, we will obtain structural information in the system by deep learning and improve the stability of structural line detection. Also, the planar constraints will be introduced for mapping.

\section*{ACKNOWLEDGMENT}
This work was supported by the Foundation for Innovative Research Groups at the National Natural Science Foundation of China (Grant No. 41721003) and the fellowship of China National Postdoctoral Program for Innovative Talents (Grant No. BX20200251)

	\balance
	
	\bibliographystyle{IEEEtran}
	\bibliography{mybibfile}
\end{document}